\documentclass[10pt,twocolumn,letterpaper]{article}

\usepackage{isba}
\usepackage{times}
\usepackage{epsfig}
\usepackage{graphicx}
\usepackage{amsmath}
\usepackage{amssymb}
\usepackage[table]{xcolor}
\usepackage{array}
\usepackage{url}
\newcolumntype{M}[1]{>{\centering\arraybackslash}m{#1}}

\isbafinalcopy 

\ifisbafinal\fi
\begin{document}

\title{VGR-Net: A View Invariant Gait Recognition Network}

\author{Daksh Thapar, Aditya Nigam\\
Indian Institute of Technology Mandi\\
Mandi, India\\
{\tt\small s16007@students.iitmandi.ac.in, aditya@iitmandi.ac.in}
\and
Divyansh Aggarwal\\
Indian Institute of Technology Jodhpur\\
Jodhpur, India\\
{\tt\small aggarwal.1@iitj.ac.in}
\and
Punjal Agarwal\\
Malaviya National Institute of Technology Jaipur\\
Jaipur, India\\
{\tt\small agrawalpunjal@gmail.com}}

\maketitle
\thispagestyle{empty}

\begin{abstract}
Biometric identification systems have become immensely popular and important because of their high reliability and efficiency. However person identification at a distance, still remains a challenging problem. Gait can be seen as an essential biometric feature for human recognition and identification. It can be easily acquired from a distance and does not require any user cooperation thus making it suitable for surveillance. But the task of recognizing an individual using gait can be adversely affected by varying view points making this task more and more challenging. Our proposed approach tackles this problem by identifying spatio-temporal features and performing extensive experimentation and training mechanism. In this paper, we propose a 3-D Convolution Deep Neural Network for person identification using gait under multiple view. It is a 2-stage network, in which we have a classification network that initially identifies the viewing point angle. After that  another set of networks (one for each angle) has been trained to identify the person under a particular viewing angle. We have tested this network over CASIA-B publicly available database and have achieved state-of-the-art results. The proposed system is much more efficient in terms of time and space and performing better for almost all angles. 
\end{abstract}

\section{Introduction}
In current perilously dynamic scenarios, continuous human personal authentication is essentially required to handle any socially inimical activity. Unlike, traditionally used forms of identification and authentication such as passwords, ID cards, tokens $etc.$, biometric traits do not require to be memorized and cannot be shared, as they are unique to an individual. Hence they have started to replace traditional methods in the recent past. Biometrics can be divided into two classes : 1) Physiological and 2) Behavioral. The former includes traits like face, fingerprint, iris, ear, knuckle, palm $etc.$ The later includes signature, voice, gait $etc.$ Physiological traits are usually unique and highly discriminant. But these traits require cooperation from the subject along with a comprehensive controlled environmental setup, for efficient and accurate authentication. Hence, alone these traits are not very useful, especially in surveillance systems. On the other hand behavioral traits such as gait, enable us to recognize a person using standard cameras and even at a distance. Hence, the easy accessibility and low susceptibility to the noise makes gait a good biometric trait that can be used in video surveillance. Even within a controlled environment, fusing behavioral biometrics such as gait with other physiological biometrics have shown to give very promising results.

\subsection{Motivation and Challenges}
Human gait has many advantages over the conventional biometric traits (like fingerprint, ear, iris $etc.$) such as its non-invasive nature and comprehensible at a distance. Therefore, gait recognition has applications in various areas, such as authentication and detection of impostor demeanor during video surveillance, personal authentication as well as in several other security related fields. Since, gait features are chief behavioural characteristics of any person, these features are believed to be hard for circumvention and can be monitored through any surveillance system, constituting gait as an essential biometric personal characteristic. Unfortunately, gait recognition can be severely affected by many factors including viewing angle, clothes, presence of bags, surroundings illumination making it a really challenging problem. Hence, multi-view gait recognition can be seen as a major problem.  Efficient and accurate recognition from any viewing angle or any camera angle makes this problem extremely hard. 

\begin{figure*}[htp]
\begin{center}
\includegraphics[width=0.95\linewidth,height=0.2\linewidth]{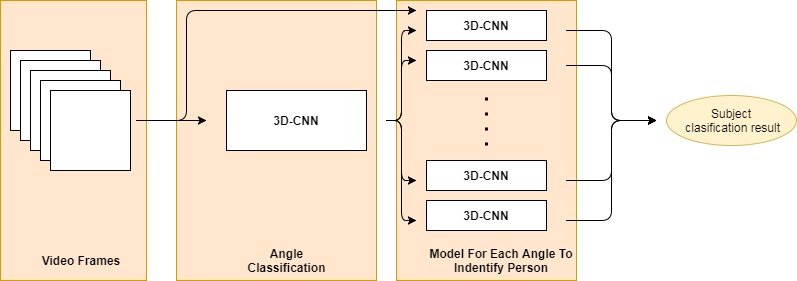}
\end{center}
\caption{Our proposed multi-view gate recognition network}
\label{fig:1}
\end{figure*}

Multi-view recognition is essentially required as most of the times, input from surveillance systems are not in accordance to the viewing angle of the enquirer resulting in very hard recognition task. In order to tackle these challenges, we propose an efficient, 3-D CNN based network for multi-view gait recognition as shown in Fig.~\ref{fig:1}. 

\subsection{Related Work}
Huge amount of work has been done in order to progress the state-of-the-art of gait recognition, started by distinguishing human locomotion from other locomotion ~\cite{1}. Till now one can classify gait feature extraction and matching broadly into two approaches : a) Handcrafted features and matching and b) Deep-Learning based approaches. 

\textbf{Handcrafted Features : } Liu et al.~\cite{liu2002gait}, used frieze patterns and combined them with dynamic time warping. This approach worked well with similar view points. Kale et al.~\cite{kale2004identification}, used the person silhouette and applied hidden markov model over it to optimize the performance. Other set of approaches used Gait Energy Image (GEI) that is an average of silhouettes over a whole gait cycle. Man et al.~\cite{man2006individual} and Hofmann and Rigoll~\cite{hofmann2013exploiting}, further  used GEI for the recognition purpose. In approaches observed in ~\cite{zheng2011robust}, ~\cite{kusakunniran2012gait} and ~\cite{muramatsu2015gait},  transformations of gait sequences into a particular desired view point has been performed. 

\textbf{Deep Learning : } Recently researchers have started to exploit 3D features for multi-view gait recognition advancing problems from 2D image classification to 3D video classification. Karpathy et al.~\cite{2} use a multi-resolution, foveal architecture by applying 3D convolutions on different time frames of a video. Similar to ~\cite{2}, Tran et al. ~\cite{tran2015learning} have designed a CNN using 3D convolutions with a deeper structure, that can fully exploit spatio-temporal features useful for video classification. Deqing et al.~\cite{4} showed us how the optical flow can be very informative for classification and identification in videos. Thomas et al.~\cite{5} uses 3D convolutional neural networks with a three channel input, with gray-scale, optical flow in X direction and optical flow in Y direction considered as three input channels. They also demonstrated that  how, playing with the training and testing data can give us better results. Shiqi Yu et al.~\cite{6}, showed that how one can use GAN networks to correctly generate GEI images (of other views) and further learn a person's gait features. Yang et al.~\cite{7}, have used  LSTM networks to memorize and extract the walking pattern of a person by correctly extracting the heat-map information.

\subsection{Contribution } 

In this work we have used only silhouettes of persons instead of other useful but expensive optical flow or GEI information. The proposed network has been trained on less amount of data without any overlapping. This also resulted in faster training and testing of network. In order to achieve view point invariant recognition, a multi-stage network initally performing viewing angle classification and later fine level subject classification. Multi-level classification has been performed by using voting at clip level. Finally we also have used stereo image data representation to train our network in order to enhance our  results.

 The remaining paper has been organized as follows, next section describes the proposed architecture. Section 3,  illustrates the experimental analysis and last section concludes the proposed work.


\begin{figure*}[htp]
\begin{center}
\includegraphics[width=0.95\linewidth, height=0.8\linewidth]{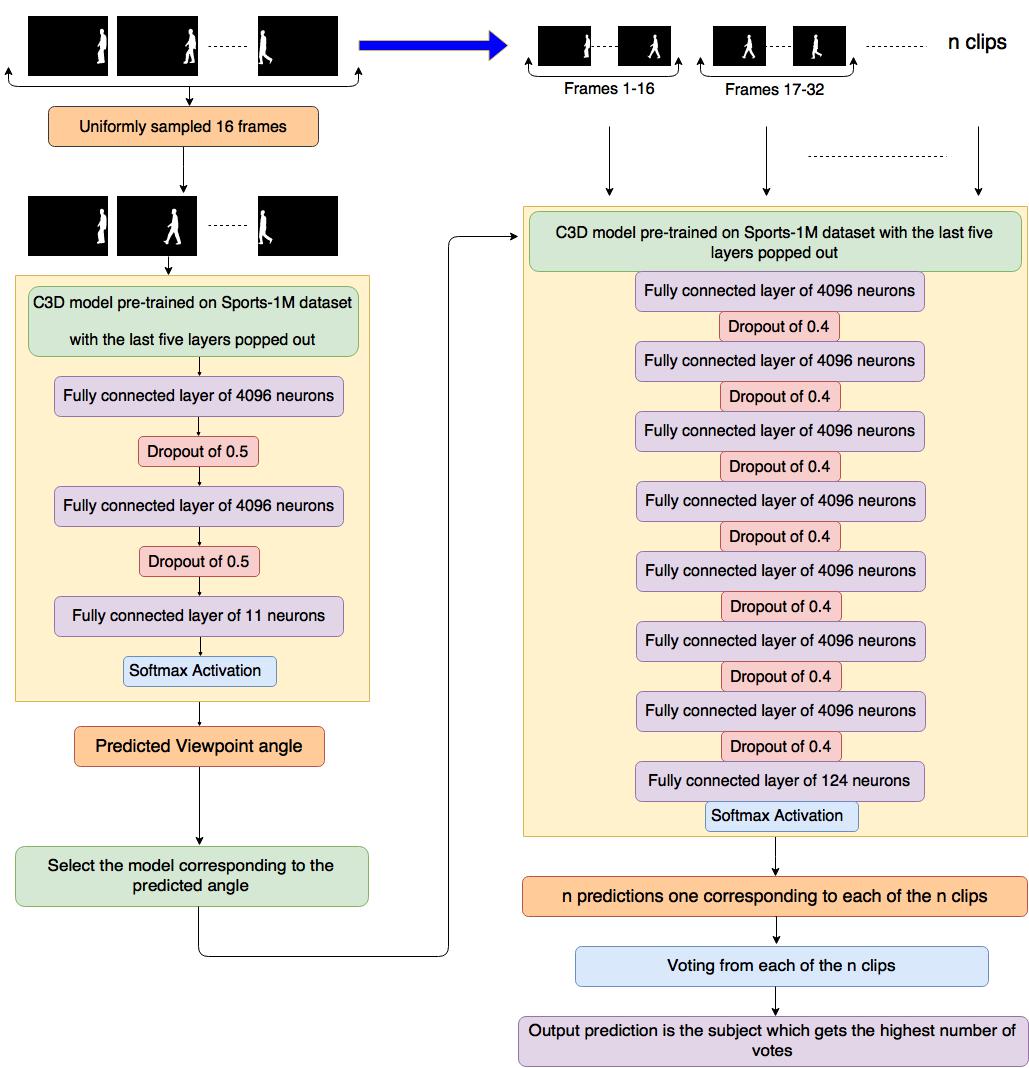}
\end{center}
\caption{Hierarchically Generalized Multi-View Gait Recognition Architecture}
\label{fig:2}
\end{figure*}

\section{Proposed Architecture}
In this work we have utilized gait video sequences for recognition and that too at various angles. Hence we have considered spatio-temporal features. After analyzing the temporal domain, one can extract a lot of features obtained from surroundings using CNN's, that can be used, further for video classification. Our proposed approach is a hierarchical two step process. In both steps deep 3D CNN's have been trained to learn the spatio-temporal features, among the video frames. The network details are shown in Fig.~\ref{fig:2}. The first network attempts  to identify and learn to estimate the viewing angle and then it attempts to perform subject identification. In order to realize this network, we have started with a basic 3D-convolutional network architecture and added few dense and pooling layers. Later we have fine tuned our network by optimizing it in terms of performance.

\subsection{Spatio-Temporal Features}
The 2D Convolutional neural networks can only analyze and learn the spatial features present in the images. In order to learn temporal gait features we need to explore another temporal dimension using 3D CNN. The spatio-temporal features relate space and time together (having both spatial extension and temporal duration).

\subsection{C3D : 3D CNN Architecture}


C3D (Convolutional 3D)~\cite{tran2015learning}, ~\cite{jia2014caffe}, ~\cite{2} is a network based on 3D convolutions which has achieved state-of-the-art results in action recognition and scene classification. It is established that when 2D convolutions are applied over successive frames of a video, the temporal information present in the frames is lost. Thus 3D convolutions and pooling are necessary to retain the time variant information in the video. 

\textbf{[a]} The network takes a $16$ frames clip each of size $112\times112$ as an  input. It contains $8$ convolution layers, $5$ pooling layers, $2$ fully connected layers followed by a softmax classification layer. Each of the convolution layers has ReLU activation applied over it with padding to preserve the spatial dimensions. 

\textbf{[b]} The first $Conv$ layer consists of $64$ filters each of size $3\times3\times3$ followed by a pooling layer of $2\times2\times1$ with a stride of $2\times2\times1$ where $1$ corresponds to the temporal domain. Then another $Conv$ layer of $128$ filters each of size $3\times3\times3$ has been applied followed by a pooling layer of pool size $2\times2\times2$ and stride of $2\times2\times2$. 

\textbf{[c]} After these initial layers two similar $Conv$ layers, each consisting of $256$ filters each of size $3\times3\times3$ followed by a max pooling of pool size $2\times2\times2$ and stride $2\times2\times2$ have been introduced. 

\textbf{[d]} Now two sets of layers are stacked after this each with two $Cov$ followed by a  pooling layer. All $Conv$ layer have to learn $512$ filters, each of kernel size $3\times3\times3$, followed by a pooling layer of $2\times2\times2$ with a stride of $2\times2\times2$. 

\textbf{[e]} Finally the output is flattened and two fully connected layers each of $4096$ neurons have been added with a dropout of $0.5$ to handle over-fitting. At last a dense layer of $487$ has been applied with a softmax activation to classify the video. 

This model has been pre-trained on the Sports-1M  dataset \cite{sp1}, which is one of the largest video classification benchmarks and contains $1.1$ million sports videos.

\subsection{Model Architecture}

\begin{enumerate}
\item \textbf{Stage-1 : }
The first stage of the proposed network uses $3D$ convolutions, to identify the viewing angle of any gait video. Sixteen frames are uniformly sampled from the video and replicated into three channels to pass through the network. The proposed network consists of a slightly tweaked version of C3D model which was already trained on Sports-1M data-set and takes as input $16$ frames each of size $112\times112$, as shown in left side of Fig.~\ref{fig:2} and discussed below. 

\textbf{[i]} The last $5$ layers of the C3D model have been removed as we need only the basic temporal features and not the action specific features ($i.e.$ \textbf {\cite{tl}}). 

\textbf{[ii]} The C3D flatten layer output gives a feature vector of size $8192\times1$. Later we have added a fully connected layer of $4096$ neurons to the above output and apply $ReLU$ activation to achieve non-linear sparsity. A dropout of $0.5$ has again been applied to avoid over-fitting and force neurons to learn generic features instead of the features specific to a given training data. 

\textbf{[iii]} A fully connected layer of $4096$ neurons is applied as a second last layer over the output with ReLU activation and a dropout of $0.5$, so as to achieve the best possible latent representation. 

\textbf{[iv]} Finally output layer of $11$ neurons has been added to classify any given videos into $11$ different viewing angles ranging from $0^\circ$ to $180^\circ$ at a difference of 18 degrees. A Softmax activation is applied over the output to get the probabilities corresponding to each of the viewing angles, since it is a classification problem.

 \begin{table}[htp]
 	\begin{center}
 		\begin{tabular}{|p{2cm}|p{2.8cm}|p{2.8cm}|}
 			\hline
 			\multicolumn{2}{|c|}{\textbf{Parameter}} & Value\\
 			\hline\hline
 			\multicolumn{2}{|c|}{\textbf{Optimizer}}  & Stochastic Gradient Descent optimizer\\
 			\hline
 			\multicolumn{2}{|c|}{\textbf{Epochs}} & 90\\
 			\hline
 			\textbf{Dual Learning Rate}  &(For 1\textsuperscript{st} 40 epochs)& 0.005 \\
 			
 			 &(For rest 50 epochs)& 0.003 \\
 			\hline
 		    \multicolumn{2}{|c|}{\textbf{Momentum}} & $0.90$ \\
 		    \hline
 			\multicolumn{2}{|c|}{\textbf{Mini batch size}} & 16\\
 			\hline
 		\end{tabular}
   	\end{center}
 	\caption{Summarizing the Person Identification Model Parameters}\label{table:1}
 \end{table}

\item \textbf{Stage-2 : }
In the second stage for each of the $11$ different viewing angles a separate network has been trained only on the videos of that angle corresponding to all the subjects. This network is also inspired by the C3D model trained on the Sports-1M data-set, as shown in right side of Fig.~\ref{fig:2} and discussed below.

\textbf{[i]} The last five layers of the network has been removed and the output of the flatten layer of the model is taken. 

\textbf{[ii]} Seven blocks, each consisting of a fully connected layer of $4096$ neurons with a ReLU activation and a dropout of $0.4$ are stacked sequentially after the output of the flatten layer. 

\textbf{[iii]} Finally a fully connected layer of $124$ neurons is added with Softmax activation to get the class probabilities for each of $124$ subjects. Since $16$ frames can be fed into the C3D model at a time, original video is cut into clips of $16$ frames each. 

\textbf{Voting Scheme : } While training each of these clips has been considered as a different video and error is calculated and back-propagated by comparing it with the ground-truth of each clip. However during testing the trained network has been used to predict the subject corresponding to each of the clips independently under a voting scheme. The subject which receives the highest number of votes from the clips has been referred as the final prediction.

\end{enumerate}

\begin{table}[htp]
 	\begin{center}
 		\begin{tabular}{|p{2.2cm}|p{2.85cm}|p{2.8cm}|}
 			\hline
 			\multicolumn{2}{|c|}{\textbf{Parameter}} & Value\\
 			\hline\hline
 			\multicolumn{2}{|c|}{\textbf{Optimizer}}  & Stochastic Gradient Descent optimizer\\
 			\hline
 			\multicolumn{2}{|c|}{\textbf{Epochs}} & 115\\
 			\hline
 			
 			\textbf{Multiple Learning Rate}  &(For 1\textsuperscript{st} 25 epochs)& 0.001 \\
 			
 			 &(For next 40 epochs) &0.005 \\
 			 
 			&(For last 50 epochs) & 0.003\\
 			\hline
 		    \multicolumn{2}{|c|}{\textbf{Momentum}} & $0.92$ \\
 		    \hline
 			\multicolumn{2}{|c|}{\textbf{Mini batch size}} & 16\\
 			\hline
 		\end{tabular}
   	\end{center}
 	\caption{Summarizing the Model Parameters for Stereo Training}\label{table:4}
 \end{table}
 
\section{Experimental Analysis}
We have conducted experiments on CASIA-B dataset~\cite{casiab} to show the utility of our model. The Correct Classification Rate (CCR\%) is computed for performance evaluation. The results are computed under two scenarios namely (a) Stage-1 classification : Predicting viewpoint angle and (b) Stage-2 classification :  Personal identification. In this work we have compared our 2-stage deep 3D convolutional neural network with the current state-of-the-art network proposed in ~\cite{5}. We have successfully demonstrated that the proposed network is efficient in terms of time and space as well as outperforms \cite{5} in terms of performance (i.e. CCR\%).

\begin{figure}[htp]
	\begin{center}
	\includegraphics[width=0.98\linewidth, height = 0.6\linewidth]{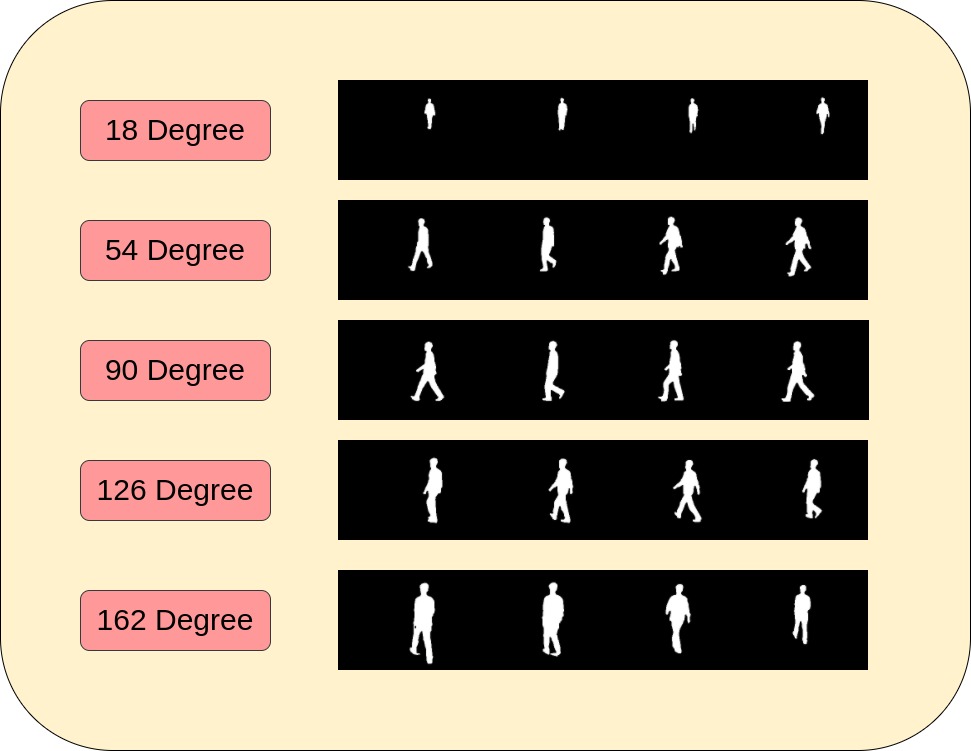}
	\end{center}
	\caption{Sample silhouettes of a subject at five viewing angles. }
	\label{fig:3}
\end{figure}

\subsection{Database Specification}
The CASIA-B Dataset \cite{casiab}, is a large multi-view gait database. This dataset consist of $102,828$ gait images corresponding to $124$ subjects, and the gait data has been captured from $11$ views ranging from $0^{\circ}$ to $180^{\circ}$. There are six sequences of normal walking corresponding to each of the $124$ subjects and for all $11$ different angles. Instead of the original video, we have used human silhouettes extracted from the videos for our experiments as shown in Fig.~\ref{fig:3}.

\subsection{Performance Parameters}
We have used the correct classification rate (CCR) as our performance parameter. It gives us an idea about how many times we have correctly identified the view point angle or  the person in a video.

\subsection{Training and Testing Protocol}
For angle identification network, we have used $82$ subjects for training that corresponds to $66\%$ of the total data whereas the remaining $42$ subject videos have been used for testing corresponding to $34\%$ of the total dataset.
For person identification network, we have used first four sequences from each subject for training whereas two sequences have been used for testing.

\begin{figure}[htp]
	\begin{center}
	\includegraphics[width=0.98\linewidth]{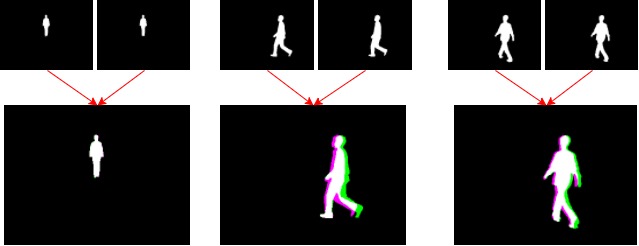}
	\end{center}
	\caption{Formation of stereo images using silhouettes}
	\label{fig:ste}
\end{figure}
\textbf{Training hyper-parameters and strategy : } Person identification networks (Stage-2), have been trained by using stochastic gradient descent $(sgd)$ optimizer with a learning rate of $0.005$ for 1\textsuperscript{st} $40$ epochs. Later we have decreased it to $0.003$ for next $50$ epochs so as to utilize dual diminished learning rate for fast convergence and accuracy but with a fixed momentum at $0.9$. All model parameters are shown in Table~\ref{table:1}. After training our networks over the gray-scale silhouettes, we fine tuned it over stereo images. They are created by stacking one image over the other and finding the difference at the pixel level of the two images as shown in Fig. \ref{fig:ste}. Hence the two consecutive video frames can be fed one by one to create a fused stereo image. Over them,  we have fine tuned the model that has already been trained on silhouettes with a learning rate of $0.001$ with momentum $0.92$ for $25$ epochs. Finally we have trained the model for first $40$ epochs at $0.005$ learning rate and next $50$ epochs at a learning rate of $0.003$. Those model parameters for stereo training are shown in Table~\ref{table:4}.

\subsection{Analysis }
In this subsection we have presented a detailed and rigorous multi-stage as well as stereo/partial overlapping performance analysis. 

\subsubsection{Stage-1 : View Point Angle Identification}

\begin{figure*}[htp]
	\begin{center}
	\includegraphics[width=0.75\linewidth]{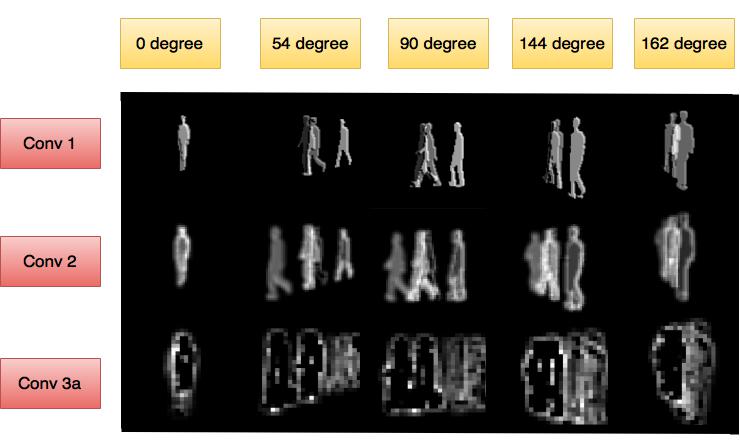}
	\end{center}
	\caption{Visualizations of various convolution layers for different angles, for proposed angle detection network}
	\label{fig:4}
	\end{figure*}

The training and testing at stage-1 has been performed on images acquired from same type of sensor model. Our model has achieved the perfect $CCR=100\%$ for angle identification. It can be clearly observed from Fig. \ref{fig:4}, that the stage-1 network has learned the underlying latent features representation successfully at different layers and hierarchy. The perfect angle identification can be easily justified by the discriminative spatio-temporal representations as shown in Fig. \ref{fig:4}. We believe that this fact attributes most significantly towards the efficiency and accuracy of our proposed overall multi-stage network.

\begin{table}[ht]
\begin{center}

\begin{tabular}{|M{1.3cm}|M{1.7cm}|M{1.5cm}|M{2.25cm}|}
\hline
\textbf{Angle}  & \textbf{Thomas~\cite{5}} & \textbf{VGR-Net} & \textbf{VGR-Net}    \\ 
 & & &(Stereo images)\\\hline \hline
0$^{\circ}$                         & 96.30\%           & 98.33\% &         \cellcolor{green}98.67\%  \\ \hline
18$^{\circ}$                        & 98.20\%           & 99.17\% &         \cellcolor{green}99.55\%\\ \hline
36$^{\circ}$                        & 98.50\%           & 99.17\% &         \cellcolor{green}100\%\\ \hline
54$^{\circ}$                        & 95.40\%           & 96.67\% &         \cellcolor{green}99.55\%\\ \hline
72$^{\circ}$                        & 94.30\%           & \cellcolor{green}97.92\% &         97.78\%\\ \hline
90$^{\circ}$                        & \cellcolor{green}99.90\%           & 97.08\% &         97.79\%\\ \hline
108$^{\circ}$                       & 98.60\%           & 97.91\% &         \cellcolor{green}98.67\%\\ \hline
126$^{\circ}$                       & 97\%              & \cellcolor{green}97.08\% &         96.90\%\\ \hline
144$^{\circ}$                       & \cellcolor{green}97.40\%           & 96.25\% &             96.38\%\\ \hline
162$^{\circ}$                       & \cellcolor{green}99.20\%           & 96.67\% &             96.10\%\\ \hline
180$^{\circ}$                       & 96.10\%           & 97.08\% &             \cellcolor{green}97.69\%\\ \hline
\end{tabular}
\end{center}
\caption{The Comparative Analysis of Quality Performance in CCR(\%) on proposed architecture and the previous architecture  for different View-Point Angles for Gray-Scale and Stereo Images. Green colored cell indicates highest accuracy for that angle.}
\label{table:2}
\end{table}
\subsubsection{Stage-2 : Person Identification}

In this stage, personal identifications task has been performed for a given view point angle. Table~\ref{table:2}, indicates the computed performance of the proposed architecture on CASIA-B dataset~\cite{casiab}. It has been observed that the proposed architecture performs better than state-of-the-art model \cite{5}, in $7$ out of $11$ viewpoint angles. 

\textbf{Angular performance analysis : } For obtuse angular view, we have visually observed that subject's velocity  with respect to the cameras has been much greater as compared with acute angles. That causes frame overlap between the clips (used for classification) to be very small for obtuse angles. In \cite{5}, huge amount of overlapping clips have been utilized for training/testing resulting in better performance for obtuse angles but at an expense of more time and space resources. In this work in order to optimize the time and space we have used non-overlapping clips apart from multi-stage network. Such a strategy enable us to achieve better performance for acute angle as well as some obtuse angles.


\begin{figure*}[htp]
	\begin{center}
	\includegraphics[width=0.95\linewidth, height=0.6\linewidth]{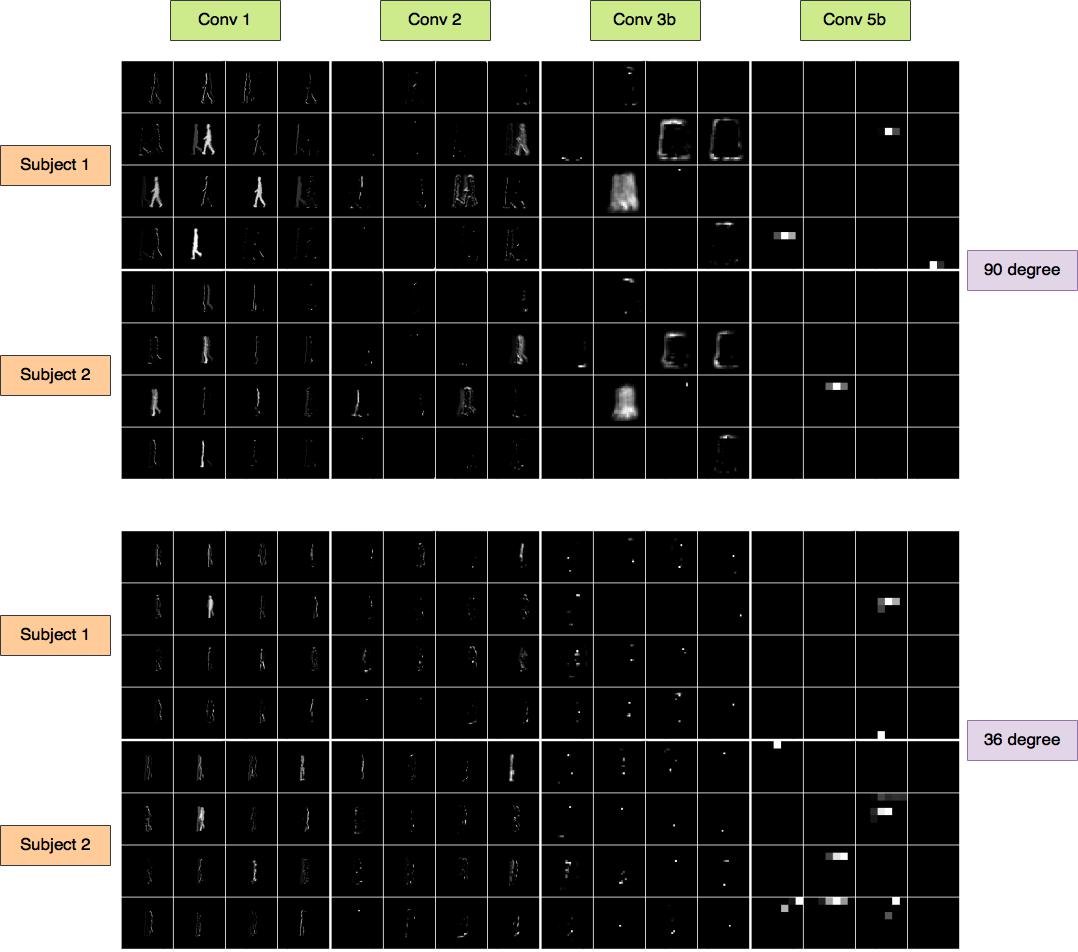}
	\end{center}
	\caption{Visualizations of the second model for two different subjects at 90 degree and 36 degree }
	\label{fig:5}
	\end{figure*}
	
\textbf{Experimentation with stereo images : } Since they have used optical flow also for the prediction, we also have tested our model on stereo images. The $CCR\%$ has been shown in Table~\ref{table:2}, which shows that using stereo images instead of normal silhouettes will lead to higher accuracy specially for acute angles primarily due to above mentioned reason.

\begin{table}[ht]
\begin{center}
\begin{tabular}{|M{1.2cm}|M{1.7cm}|M{1.5cm}|M{2.35cm}|}
\hline
\textbf{Angle}& \textbf{Thomas~\cite{5}} & \textbf{VGR-Net} & \textbf{VGR-Net}    \\ 
 & & &(partial overlap)\\\hline \hline
90$^{\circ}$                       & \cellcolor{green}99.90\%           & 97.08\%          &98.55\%\\ \hline
108$^{\circ}$                       & 98.60\%           & 97.91\%          &\cellcolor{green}98.75\%\\ \hline
144$^{\circ}$                       & 97.40\%           & 96.25\%          &\cellcolor{green}{97.50\%}\\ \hline
162$^{\circ}$                       & \cellcolor{green}99.20\%           & 96.67\%          &98.33\%\\ \hline

\end{tabular}
\end{center}
\caption{The Comparative Analysis of Quality Performance in CCR(\%) on proposed architecture and the previous architecture  for different View-Point Angles for Gray-Scale and with partial overlap. Green colored cell indicates highest accuracy for that angle.}
\label{table:3}
\end{table}

\textbf{Experiment with ``partial overlap'' : } 
It is well understood that overlapping  clips can handle the problem adequately but at an expense of more time and space. Hence we have also tested our network with a minimal overlap (almost 50\% as considered in \cite{5}). In our model overlapping has been done as follows : 1 to 16 frames in 1 \textsuperscript{st} clip then  8 to 24 frames in 2\textsuperscript{nd} clip and so on on contrary to~\cite{5}, where they took complete overlap between the clips resulting them a very slow training and testing network. We have tested the model with ``partial overlap'' only at 90, 108, 144 and 162 degrees to justify the above arguments.  The ``partial overlap'' results are shown in Table~\ref{table:3}. One can clearly observe that by considering ``partial overlap'', the performance of our model increases and even surpasses \cite{5} at obtuse angles. 

\subsection{Visualizations}
We have visualized the features extracted at different layers of our proposed network, since these are the actual images as seen at any layers. We have critically analyzed them, so as to understand how and what our network has learned and interpreting about any image/clip.

\textbf{Angle Identification : } The Fig.~\ref{fig:4} depicts outputs of $1^{st}$, $2^{nd}$ and $3a$ convolutional layers of the network model, at different angles, for angle identification network. Clearly it is visible that the filters have learned the spatial as well as the temporal features that can easily discriminate the angles. The first convolution layer consists of filters of size $3\times3\times3$ and they are trained to learn the motion in three adjacent video frames. Since at 0 degree, the person comes towards the camera, that appears to be just one silhouette that overlaps the previous frames silhouettes. Since till now no pooling has been done in spatial or temporal domain, the silhouette edges are clearly visible. In second convolution layer one can  notice that the filters learn deeper and much more complex and higher level features. Since pooling has been done in the spatial domain, the edges seem to fade a little and only the basic outline of the silhouettes tend to remain. However, as yet no pooling has been done in the time variant zone, therefore similar to the first $Conv$ layer, only features from adjacent three frames of the video are learned by a filter. 

From the output of this layer it is clearly visible that $0$ degree can be quiet easy to differentiate from rest of the angles because the successive silhouettes (for other angles) overlap with the existing ones and thus overshadow them, leaving the latest silhouettes visible. Output of the $3rd$ convolution layer aggregate the features in all three dimensions, as just before it a max pooling of 2 with a stride of 2 has been done. Hence filters in this layer learn features at another scale. The filters learn  motion in consecutive frames leaving all other irrelevant spatial information. Since in this network the aim is to identify the view-angle, therefore these deeper filters learn the direction of flow in the frames and lose the information about the person identity that may be present in the spatial domain.

\textbf{Personal Identification : }
In our stage-2 model (personal identification network),  we have noticed that the neurons try to learn not only the movement but also the spatial salient features in the silhouettes. This is due to the fact that, now the network needs to learn those latent features that can effectively differentiate between various subjects as shown in Fig.~\ref{fig:5}. During the initial convolution layers like $conv1$, $conv2$ and $conv3b$ the filters tend to learn local features such as the basic outline of the silhouettes and the aggregated motion in the successive silhouettes which can help the further classifiers to classify the different subjects. As we move on to the more deeper convolution layers in the network such as $Conv5b$ we have observed that the neurons tends to learn more global features. Here only those neurons which are responsible for the classification of that subject tend to fire. It can be observed in Fig.~\ref{fig:5} that only some of the neurons fire because they are the classification neurons for that particular subject. In Fig. \ref{fig:5}, two subject comparative visualization has been shown at two different angles $viz.$ $90$ and $36$ degrees. Over the former angle ($i.e$ 90 degrees) our results are slightly lesser, as the features seen by network tends to become more general and ambiguous for some subjects at complete profile angle. While sparse and unique features got extracted from the network at $36$ degrees.

\section{Conclusion}
We have proposed an efficient (in terms of time and space) and accurate (in terms of performance) architecture for multi-view gait recognition in comparison to the  present state-of-the-art network \cite{5}. Our network is viewpoint invariant due to its multi-stage architecture. We have not considered complete overlapping clips as well as any other pre-computed optical-flow features as they need to be computed outside the network and slow down the network significantly. Instead we have only used the basic silhouette and have decreased the running time of the whole system considerably as well as achieved better/comparable results to the present state-of-the-art system. In order to show that overlapping and optical flow can easily boost the system performance, we have also utilized  stereo images as well as small overlapping clips for better performance but at an expense of additional cost.

{\small
\bibliographystyle{ieee}
\bibliography{submission_example}
}

\end{document}